\documentclass[letterpaper, 10 pt, conference]{ieeeconf}  % Comment this line out if you need a4paper

\IEEEoverridecommandlockouts                              % This command is only needed if 
\usepackage[table,xcdraw]{xcolor}
\usepackage{verbatim}
\usepackage{amsmath}
\usepackage{amsfonts}
\usepackage{graphicx}
\usepackage{array}
\usepackage{float}
\usepackage{makecell} % 用于单元格内容排版
\usepackage{multirow} % 用于跨行操作
\usepackage{booktabs} % 用于美观的表格横线
\usepackage{caption} % 用于设置表格标题等样式
\usepackage{cite}
\usepackage{caption}
\usepackage{tocloft}
\usepackage{subcaption}

\title{\LARGE \bf
KiteRunner: Language-Driven Cooperative Local-Global Navigation Policy with UAV Mapping in Outdoor Environments
}
\author{Shibo Huang$^{1*}$, Chenfan Shi$^{1*}$, Jian Yang$^{2\dagger}$, 
Hanlin Dong$^{1}$, 
Jinpeng Mi$^{3}$,  Ke Li$^{2}$, Jianfeng Zhang$^{1}$,\\ 
 Miao Ding$^{4}$, Peidong Liang$^{5}$, 
 Xiong You$^{2}$, 
Xian Wei$^{1\dagger}$
\thanks{
$^{*}$Equal technical contribution;
$^{1}$Software Engineering Institute, East China Normal University;
$^{2}$School of Geospatial Information, Information Engineering University; 
$^{3}$Institute of Machine Intelligence (IMI), University of Shanghai for Science and Technology;
$^{4}$School of Software, Liaoning Technical University;
$^{5}$Fujian (Quanzhou) Institute of Advanced Manufacturing Technology;
$^{\dagger}$Corresponding Author.}
}

\begin{document}

\maketitle
\thispagestyle{empty}
\pagestyle{empty}

%%%%%%%%%%%%%%%%%%%%%%%%%%%%%%%%%%%%%%%%%%%%%%%%%%%%%%%%%%%%%%%%%%%%%%%%%%%%%%%%
\begin{abstract}
Autonomous navigation in open-world outdoor environments faces challenges in integrating dynamic conditions, long-distance spatial reasoning, and semantic understanding. Traditional methods struggle to balance local planning, global planning, and semantic task execution, while existing large language models (LLMs) enhance semantic comprehension but lack spatial reasoning capabilities. Although diffusion models excel in local optimization, they fall short in large-scale long-distance navigation. To address these gaps, this paper proposes KiteRunner, a language-driven cooperative local-global navigation strategy that combines UAV orthophoto-based global planning with diffusion model-driven local path generation for long-distance navigation in open-world scenarios. Our method innovatively leverages real-time UAV orthophotography to construct a global probability map, providing traversability guidance for the local planner, while integrating large models like CLIP and GPT to interpret natural language instructions. Experiments demonstrate that KiteRunner achieves $5.6\%$ and $12.8\%$ improvements in path efficiency over state-of-the-art methods in structured and unstructured environments, respectively, with significant reductions in human interventions and execution time.
\end{abstract}
% \begin{abstract}
% This paper introduces a novel language-driven robotic navigation policy, termed the Language-Driven Cooperative Local-Global Navigation Policy (LaCoLG-Nav).
% The method integrates local-global planning strategies to ensure robust and efficient autonomous navigation in unstructured outdoor environments. 
% It combines a flexible local path planner with UAV orthophoto-based global hints for efficient and reliable navigation performance.
% The local planner generates adaptive paths in real-time using a diffusion model, while the global planner leverages traversability identification from the large-scale map to achieve long-range navigation.
% Additionally, we introduce language-driven navigation, utilizing large-scale pre-trained vision-language models, including CLIP for visual-language understanding and GPT for processing complex user commands. This integration enables the robot to interpret and respond to high-level tasks naturally, improving human-machine interaction. 
% We evaluated the performance of LaCoLG-Nav in both structured and unstructured environments, demonstrating that the proposed method achieves improvements in average path efficiency of 5.6\% and 12.8\% compared to the most advanced method, respectively.
% %Extensive experimental results demonstrate that LaCoLG-Nav significantly improves navigation efficiency, robustness, and adaptability in complex dynamic environments, highlighting its strong performance in real-world task execution.
% \end{abstract}

%%%%%%%%%%%%%%%%%%%%%%%%%%%%%%%%%%%%%%%%%%%%%%%%%%%%%%%%%%%%%%%%%%%%%%%%%%%%%%%%
\section{INTRODUCTION}
\begin{figure*}[thpb]
  \centering
  \includegraphics[scale=0.085]{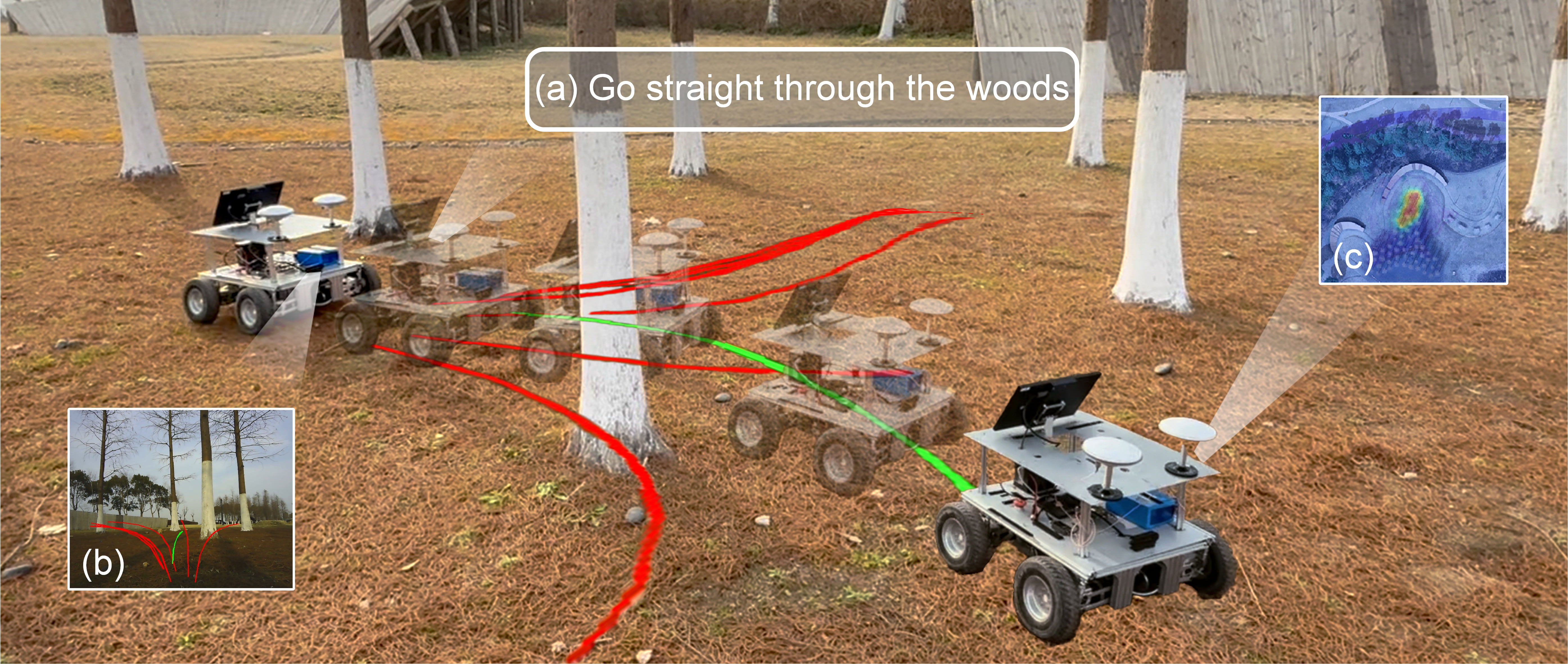}
  \caption{KiteRunner accepts natural language instructions and realizes long-distance outdoor navigation through the cooperation of local planner and global planner. (a) natural language commands, (b) the Local Planner generating context-aware candidate paths, and (c) the Global Planner providing traversability guidance.}
  \label{xiaoguo}
\vspace{-7pt} %缩减空白
\end{figure*}

%卫星的实时性差；
%户外大场景导航很难
Autonomous navigation in unstructured open-world outdoor environments remains a significant challenge in robotics, primarily due to the dynamic and unpredictable nature of large-scale outdoor environments.
%需要全局局部结合以及语义理解
This requires a robot system that can take into account both local planning and global planning capabilities, while having a comprehensive semantic understanding of outdoor open-world environments and a high-level understanding of its own tasks\cite{guan2022ga}.
% high-level task understanding

%传统方法例如。。。。无法很好的完成这一任务
Traditional approaches, including graph-based algorithms such as A* and Dijkstra\cite{ruan2014map,9391698}, deep learning based methods\cite{tai2017virtual, wang2024autonomous} and reinforcement learning-based methods\cite{zhu2021deep,devo2020deep,lu2021mgrl}, are difficult to conduct outdoor long-distance exploration, since their training scenarios cannot cover the variability of the real open-world outdoor environments. They also lack the semantic understanding and reasoning ability of the open world, making it difficult to perform complex tasks.

%大模型增强语义理解
To address this problem, recent progresses have been achieved by leveraging the strong generalization capabilities of large deep learning models (e.g., CLIP\cite{radford2021learning} and GPT\cite{zhou2024navgpt}) in scene recognition and semantic understanding\cite{lopez2020semantic}. 
The methods based on Large Language Model (LLM), such as LM-Nav\cite{shah2022lmnav} and BehAV\cite{weerakoon2024behav}, leverage pre-trained LLMs to interpret natural language instructions and ground them in visual observations, enabling robotic navigation without extensive task-specific fine-tuning.
Although these methods enhance the semantic understanding of the large-scale outdoor environment, they are weak in spatial reasoning and have difficulties in adapting to high local environmental variability.

%基于扩散模型的方法在本地规划上表现很好，但缺乏全局能力，不足以应对室外大场景长距离导航任务
In order to adapt to the local environmental variability, diffusion model-based policy demonstrates significant advantages. For example, NoMaD\cite{sridhar2024nomad} uses diffusion models to generate multiple action sequences and selects optimal strategies through denoising processes to achieve efficient local planning and obstacle avoidance; ViNT\cite{shah2023vint} uses the diffusion model to provide diverse target suggestions for exploration tasks, helping robots efficiently search for targets. 
However, the diffusion model is good at local path optimization but lacks explicit modeling ability for long-distance global path planning, which is insufficient for long-distance navigation tasks in large-scale outdoor scenes. % which is insufficient for long-distance navigation tasks in outdoor large scenes 要缩短篇幅可删

%为了提上长距离导航能力，一些方法卫星地图，但卫星地图有xxx不足，无人机正射影像可以解决这些问题
In order to enhance the model's long-distance spatial reasoning ability, the overhead satellite map has been used in methods such as ViKiNG\cite{shah2022viking}.
However, the satellite map is not real-time and updates slowly, making it difficult to adapt to real-time changes in the environment. In addition, the top view of the satellite is oblique photography, which has problems such as shadows and occlusions. The positioning and navigation accuracy in these areas is low.
% Drone photography can be taken in real time, covering the target area and all angles, with great convenience and higher accuracy. 
UAV photography enables real-time, high-coverage imaging of target areas from multiple angles, offering superior operational flexibility and precision.
%UAV orthophoto is a good way to replace satellite maps. 
UAV orthophotos are a great alternative to satellite maps.
The digital orthophoto map (DOM) based on UAV images\cite{liu2018generating,zhang2023aerial} provide high-resolution real-time imaging with minimal occlusions compared to satellite maps.

Overall, the aforementioned methods have challenges in taking into account the local environmental variability, the long-distance spatial reasoning, and the comprehensive semantic understanding of the open-world outdoor environments at the same time.
To address these challenges, in this work, we propose a novel language-driven cooperative local-global navigation policy with UAV mapping, namely \textit{KiteRunner}, which aims to bridge these gaps by seamlessly integrating three critical components: UAV orthophoto-based global planning, diffusion models for local path planning and language-driven navigation.
By leveraging large-scale pre-trained models such as CLIP and GPT, the proposed method allows robots to interpret high-level human commands and navigate complex environments in real time. The integration of these models ensures both flexibility and precision in executing tasks, making it a significant step forward in language-grounded navigation.
Experimental results across structured and unstructured environments demonstrate that KiteRunner achieves superior performance in path efficiency, intervention count, and execution time, underscoring its robustness and adaptability to dynamic conditions.

The key contributions of this paper are as follows:

\begin{itemize}
    \item We propose a novel language-driven robotic navigation policy that combines the diffusion-based local planning model with UAV orthophoto-based global planning strategies, namely \textit{KiteRunner}. In the long-distance outdoor large-scale scene navigation experiments, our method increases the path efficiency by $5.6\%$ in structured scenes and $12.8\%$ in unstructured scenes compared to the most advanced methods.
    
    \item We propose a method to generate a global probability map based on overhead UAV orthophoto maps, instead of satellite maps, to provide robots with global traversability information in real time. 

    \item We propose a localization-aware local planner that enables robots to rapidly determine their positions within topology maps during long-distance navigation tasks, while ensuring minimal deviation from global paths.
    % \item We propose a local planner with global information hints to help robots achieve long-distance navigation in large outdoor scenes.
    
%    Allow robots to have a higher ability to adapt to local environmental changes
    
    % \item We propose a method of using UAV digital orthographic maps instead of traditional satellite maps to make the overhead map more real-time and precise.
    
    % \item We provide extensive experimental validation, showing significant improvements in navigation efficiency, robustness, and adaptability in diverse real-world environments.
\end{itemize}
%本文的主要贡献如下：
%1. 我们提出了一种新颖的语言驱动的机器人导航方案，将局部扩散模型与基于无人机的全球规划策略相结合。
%2.我们展示了由基于无人机的高架地图导出的全球概率图如何指导路径选择，增强自主导航中的决策。
%3.我们介绍了语言驱动的导航功能，其中预训练的视觉语言模型 允许机器人解释和执行高级用户命令，提高了灵活性和适应性。 我们提供了广泛的实验验证，显示了在不同现实世界环境中导航效率、鲁棒性和适应性的显着提高。
\vspace{-5pt} %缩减空白

\section{Related Work}
Robust and efficient autonomous navigation in unstructured outdoor environments has been a critical challenge in robotics. Existing approaches have addressed various aspects of navigation, including local path planning, global navigation, and semantic reasoning. This section reviews recent advancements in these areas, with a focus on language-driven navigation, UAV orthophoto-based global path planning, and local path generation models.
%非结构化室外环境中鲁棒高效的自主导航一直是机器人技术的关键挑战。现有方法已经解决了导航的各个方面，包括局部路径规划、全局导航和语义推理。本节回顾了这些领域的最新进展，重点是语言驱动的导航、基于无人机正射影像的全局路径规划、局部路径优化的扩散模型以及结合这些技术的集成系统。

\subsection{Vision-Language Integration for Semantic Navigation}
Recent advancements in visual language models (VLMs)\cite{2024arXiv240601584C,WOS:001169499004044} have significantly enhanced the capability of robots to perform complex tasks based on natural language instructions. LM-Nav utilizes pre-trained models to interpret language instructions and generate actionable navigation goals. Additionally, the models, such as CLIP\cite{radford2021learning}, that combine contrastive language-image pretraining, enable robots to understand high-level tasks by associating visual cues with textual descriptions. However, a major limitation of these models is their inability to directly handle geometric and spatial challenges, which are essential for effective real-world navigation.

Recent advancements\cite{xu2023vision,2024arXiv240920445E} aim to address these challenges by combining semantic reasoning with geometric path planning. Approaches that integrate vision-language models with geometric optimization frameworks are proving successful in overcoming these limitations. These models translate high-level commands into spatially feasible navigation plans, improving their ability to perform in dynamic, unstructured environments.
%视觉语言模型（VLM）的最新进展显著增强了机器人根据自然语言指令执行复杂任务的能力。LM-Nav等系统利用预训练模型来解释语言指令并生成可操作的导航目标。此外，CLIP等模型结合了对比语言-图像预训练，使机器人能够通过将视觉提示与文本描述相关联来理解高级任务。然而，这些模型的一个主要限制是它们无法直接处理几何和空间挑战，而几何和空间挑战对于有效的现实世界导航至关重要。

%近期的进展旨在通过将语义推理与几何路径规划相结合来应对这些挑战。将视觉语言模型与几何优化框架相结合的方法在克服这些限制方面被证明是成功的。这些模型将高级指令转化为空间上可行的导航计划，提高了它们在动态、非结构化环境中的执行能力。

\subsection{Diffusion Models for Local Path Planning}

Local path planning has evolved to focus on dynamic adaptation to obstacles and environmental changes\cite{TAO2024102254,10237217,2025arXiv250209556S}. Diffusion models have gained prominence due to their ability to generate flexible paths that adapt to real-time environmental shifts without the need for exhaustive pre-mapping\cite{wang2023diffusion,chi2023diffusion,reuss2023goal,pearce2023imitating,janner2022planning}. These models employ probabilistic decision propagation, enabling smooth navigation in environments that may not be fully mapped in advance.

Recent works like RECON\cite{shah2021rapid}, ViNT\cite{shah2023vint}, and ViKiNG\cite{shah2022viking} have extended the application of diffusion models in local path planning. RECON uses latent goal models and topological memory for efficient exploration and path planning in new environments. ViNT enhances long-range navigation by leveraging diffusion-based goal proposals to explore unfamiliar areas. ViKiNG, on the other hand, combines diffusion models with geographic hints such as satellite maps, improving long-distance navigation in previously unseen environments. Together, these approaches integrate diffusion models with topological strategies, enhancing local planners’ flexibility and adaptability in both familiar and novel environments.

Despite their advantages, diffusion models are generally limited to short-range planning. Hybrid approaches that combine local diffusion models with global path planners are emerging, addressing the need for both short-range adaptability and long-range efficiency.

\subsection{Global Planning with Overhead Map Representations}
Global path planning is critical for optimizing long-range navigation in autonomous systems\cite{Gao2019AGP,2024arXiv240701862X}. A widely adopted method in this domain is overhead map planning\cite{2024arXiv240713517C,2023arXiv230815427G}, which leverages top-down views obtained from satellite or UAV-based mapping systems to provide comprehensive spatial awareness. Although satellite-derived maps offer valuable global insights, their static nature and infrequent updates can limit effectiveness in rapidly changing environments. In contrast, UAV-generated Digital Orthophoto Map (DOM) deliver higher spatial resolution and more frequent updates, making them better suited for real-time global planning.

Furthermore, a probabilistic approach\cite{10606099, Kim2025GPT4offOT} is adopted to represent traversability. Instead of relying on binary traversable/non-traversable labels, environmental features are modeled with probability distributions. This enables smoother transitions in the planned paths and enhances adaptability to dynamic obstacles and partially observable environments\cite{zhang2025dual}.

In summary, our work presents a language-driven cooperative navigation framework that seamlessly integrates natural language processing, UAV-based global mapping, and diffusion-based local planning. This unified approach enables robust long-range guidance while maintaining agile responses to local dynamic obstacles, forming a cohesive paradigm for autonomous outdoor navigation.

\begin{figure*}[thpb]
  \centering
  \includegraphics[scale=0.9]{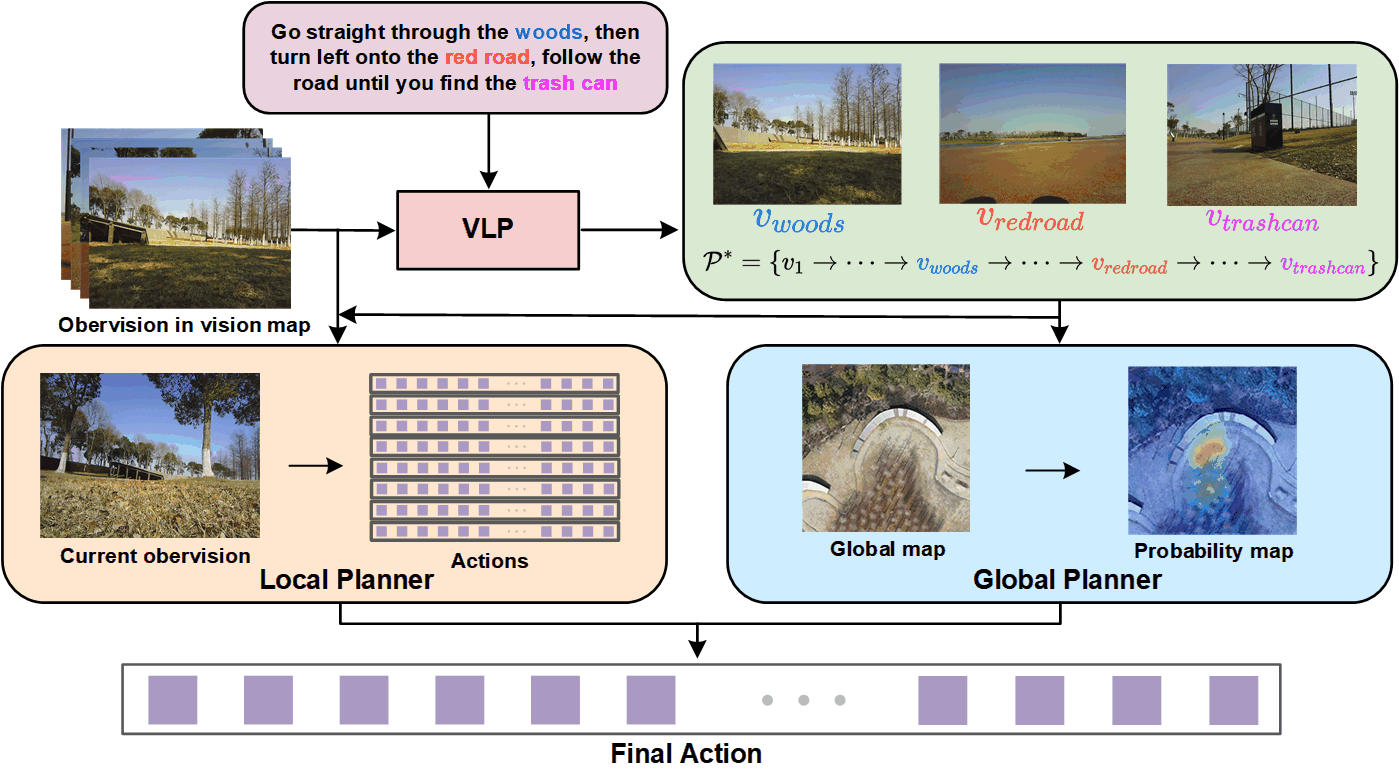}
  \caption{
  KiteRunner is suitable for outdoor navigation tasks in structured and unstructured scenes. The proposed method employs the Vision-Language Processor (VLP) to infer optimal navigation paths adhering to natural language instructions. By strategically integrating trajectory outputs from the Local Planner with traversability probability maps generated by the Global Planner, this framework achieves efficient language-guided outdoor navigation.
  %LaCoLG-Nav适用于室外结构化与非结构化场景的导航任务，该方法利用VLP推理出符合自然语言指令的最优路径，并融合了局部规划器生成的路径与全局规划器提供的可通行概率信息，从而实现高效的语言驱动室外导航能力。
  }
  \label{KiteRunner}
\vspace{-6pt} %缩减空白
\end{figure*}

\section{KiteRunner: Language-Driven Cooperative Local-Global Navigation Policy}
As illustrated in Figure \ref{KiteRunner}, the proposed KiteRunner consists of three components: (i) The Vision-Language Processor (VLP), which extracts semantic landmarks from natural language instructions and matches them with the visual landmark stored in the topology graph, generating optimal path; (ii) The Local Planner (LP), which, using the current visual input and the visual information stored in the topology graph, generates multiple candidate local paths; (iii) The Global Planner (GP), which decodes the global map to create a probability map of global traversability, guiding the LP to select the optimal path. The effective integration of these components enables robust robot navigation in outdoor environments.
%根据图1所示，所提出的KiteRunner包括三个组件：1）视觉语言处理器：它提取自然语言指令中的地标，将语义地标与拓扑图中保存的图像地标进行匹配，生成最优路径。2）局部规划器：它根据机器人当前获取的视觉输入以及拓扑图中的视觉信息，生成多条局部路径。3）全局规划器：它解码全局地图生成概率图以获取全局可通行性，从而指导局部规划器选择正确的可通行路径。这些组件的有效集成使机器人能够在户外环境中进行强大的导航。

\subsection{Vision-Language Processor:}
The VLP serves as a crucial component in KiteRunner, enabling the robot to interpret human instructions and associate them with visual landmarks in the environment. This process involves three key steps: ($1$) extracting semantic landmarks from natural language commands, ($2$) matching these landmarks in the topology graph, and ($3$) selecting the optimal navigation path based on both semantic and visual information.
%VLP在KiteRunner中充当至关重要的组成部分，使机器人能够解释人类的指示并将其与环境中的视觉地标相关联。此过程涉及三个关键步骤：（1）从自然语言命令中提取地标的描述，（2）在视觉地图中匹配这些地标，以及（3）基于语言和视觉信息选择最佳导航路径。

\textbf{Landmark Extraction from Natural Languages:} 
To parse free-form textual instructions into structured landmark descriptions, we leverage a large-scale language model, GPT-4o\cite{achiam2023gpt}, using a carefully designed prompt. The model processes user instructions and extracts an ordered list of key landmarks that guide navigation. 
% extracted landmarks are represented as:
% \begin{equation}
% L = {l_1, l_2, ..., l_n},
% \end{equation}
% where each $l_i$ corresponds to a specific environmental feature or structure (e.g., "a red bucket next to a black car").
%从自然语言中提取地标：为了将自由形式的文本说明解析为结构化的地标描述，我们使用精心设计的提示来利用大语言模型GPT-4o。该模型处理用户说明并提取指导导航的关键地标的有序列表。

% The language model is queried using:
% \begin{equation}
% L = \text{GPT-4o}(T),
% \end{equation}
% where $T$ is the input natural language text. This extraction ensures that the robot can follow high-level human commands without requiring trajectory-level annotations.
%其中T是输入的自然语言文本。这种提取确保机器人可以遵循高级人类命令，而无需轨迹级注释。

\textbf{Visual-Semantic Matching with CLIP:} Based on the extracted semantic landmark list $L$, we use the CLIP\cite{radford2021learning} to establish associations between semantic landmarks and the visual landmarks of the topology graph. For each node $v_i \in V$ in the topology graph, match its first perspective view image $\mathcal{I}_i$ with the landmark text.
%使用CLIP进行视觉语义匹配：基于提取的地标集L，我们使用 CLIP建立语义地标与拓扑图中的视觉地标之间的关联，对于拓扑图中每个节点，将其第一视角图像与地标文本进行匹配。

To quantify the visual-semantic alignment strength, we leverage CLIP's multimodal alignment capability by computing similarities between each node's visual features and all semantic landmark text features, ultimately aggregating these into the image-landmark similarity matrix image-landmark similarity matrix\cite{radford2021learning} $S \in \mathbb{R}^{|V| \times n}$, where each element $S_{i,j}$ is determined by the following equation, $\phi_{\text{vis}}(\mathcal{I}_i) \in \mathbb{R}^{d}$ is the image feature extracted by visual encoder of CLIP, $\phi_{\text{txt}}(l_j) \in \mathbb{R}^{d}$ is the landmark description feature extracted by text encoder:
%为了量化视觉 - 语义对齐强度，我们利用 CLIP 的多模态对齐能力，计算每个节点的视觉特征和所有语义地标文本特征之间的相似性，最终将其聚合到图像 - 地标相似性矩阵$S\in\mathbb{R}^{|V|\time n}$中，其中每个元素$S_{i,j}$由以下等式确定，$\phi_{\text{vis}}(\mathcal{I}_i) \in \mathbb{R}^{d}$是CLIP是视觉编码器提取的图像特征，$\phi_{\text{txt}}(l_j) \in \mathbb{R}^{d}$是文本编码器提取的地标描述特征
\vspace{-8pt} % 缩减空间

\begin{equation}
S_{i,j}= \max_{I \in \mathcal{I}_i} \left( \phi_{\text{vis}}(I) \cdot \phi_{\text{txt}}(l_j) \right)
\end{equation}

\textbf{Optimal Path Generation:} Path planning is achieved through an improved Dijkstra algorithm\cite{dijkstra2022note} that performs joint optimization of language and vision. The node value function is defined as:
\begin{equation}
Q(v) = \underbrace{\sum_{j=1}^n S_{v,j}}_{\text{semantic matching score}} - \beta \cdot \underbrace{D(v_{\text{start}},v)}_{\text{path cost}}
\end{equation}
where $\sum_{j=1}^n S_{v,j}$ represents the total semantic matching score of node \( v \) for all landmarks \( L = \{l_1, ..., l_n\} \), $D(v_{\text{start}}, v)$ is the path cost from the start node \( v_{\text{start}} \) to the current node \( v \), and $\beta$ is the weight coefficient for the path cost term, balancing the priority between semantic matching and path efficiency.
%路径规划通过对语言和视觉进行联合优化的改进 Dijkstra 算法实现，节点值函数定义为：
%$\sum_{j=1}^n S_{v,j}$反映当前节点v对所有地标的语义匹配总分,$D(v_{\text{start}}, v)$是从起点到当前节点v的路径成本，\beta是路径成本项的权重系数，用于平衡语义匹配与路径效率的优先级

% The algorithm includes a dual optimization process:
% 1. \textbf{Landmark-Driven Search:} For each landmark $l_j \in L$, the path is explored along the similarity gradient $S_{:,j}$.
% 2. \textbf{Multi-Constraint Fusion:} The results from different stages of the search are integrated via dynamic programming:
%该算法包括双重优化过程：
%1.地标驱动搜索: 对于L中的每个地标l_j\，沿着相似度梯度探索路径。
%2.多约束融合：搜索不同阶段的结果通过动态规划进行集成：

% \begin{equation}
% \mathcal{P}^* = \arg\max_{\mathcal{P}} \left[ \prod_{j=1}^n S_{v_j,l_j} \cdot e^{-\gamma L(\mathcal{P})} \right]
% \end{equation}
% where $\prod_{j=1}^n S_{p_j,j}$ reflecting the overall semantic alignment of the path with all landmarks, and $e^{-\gamma L(\mathcal{P})}$ is an exponential decay term for path length, used to penalize excessively long paths, where $L(\mathcal{P})$ is the total length of path \( \mathcal{P} \) and \( \gamma \) controls the decay rate.
%\( \prod_{j=1}^n S_{p_j,j}\)反映路径对全部地标的整体语义匹配度，e^{-\gamma L(\mathcal{P})} 是路径长度的指数衰减项，用于惩罚过长路径，其中L(\mathcal{P})是路径总长度，\gamma 控制衰减速率

This algorithm effectively balances the semantic constraints of the language instructions with the geometric constraints of the environment. The generated navigation path $\mathcal{P}^* = \{v_1 \rightarrow v_2 \rightarrow \cdots \rightarrow v_m\}$ ensures both landmark visibility and optimal movement efficiency. 
% 该算法有效地平衡了语言指令的语义约束和环境的几何约束，生成的导航路径既保证了地标的可见性，又保证了最优的移动效率。

\subsection{Local Planner}
The Local Planner (LP) operates as the real-time decision engine. In recent years, generative models have demonstrated significant potential in robotic local planning. Approaches such as ViNT generates image sub-goals through diffusion models and combine topological graph search to plan paths, while ViKiNG samples sub-goals based on latent space expand paths through geographical heuristics. However, these methods still face challenges in generating loacl paths under complex environmental constraints. Inspired by the remarkable sequence generation capabilities of diffusion models in domains such as material design, drug discovery, and chemical engineering, we pioneer their application to robotic navigation by proposing a diffusion probabilistic model-based local planner. This model generates multimodal feasible paths through an iterative denoising process, enhancing robustness to environmental uncertainties while maintaining dynamic obstacle avoidance capabilities.
%本地规划器作为实时决策引擎生成局部路径。近年来，生成式模型在机器人本地规划领域展现出巨大潜力，如ViNT通过扩散模型生成图像子目标并结合拓扑图搜索规划路径，而ViKiNG基于潜在空间采样子目标并通过地理启发式扩展路径。然而，这些方法在复杂环境下的多模态轨迹生成仍面临挑战。受扩散模型（Diffusion Model）在材料设计、药物研发等领域的序列生成能力启发，我们创新性地将其引入机器人导航领域，提出基于扩散概率模型的本地规划器。该模型通过迭代去噪过程生成多模态可行路径，在保持动态避障能力的同时，显著提升了对环境不确定性的鲁棒性。
% The Local Planner(LP) operates as the real-time decision engine, transforming perceptual inputs into local path while maintaining dynamic obstacle avoidance. 
% LP demonstrates two key innovations: 1) Diffusion-based trajectory generation, 2) Multi-modal heuristic evaluation.
%本地规划器（LP）作为实时决策引擎运行，将感知输入转换为局部路径，同时保持动态避障。LP展示了两个关键创新： 1）基于扩散的轨迹生成，2）多模态启发式评估。

\begin{figure}[thpb]
  \centering
  \includegraphics[scale=0.53]{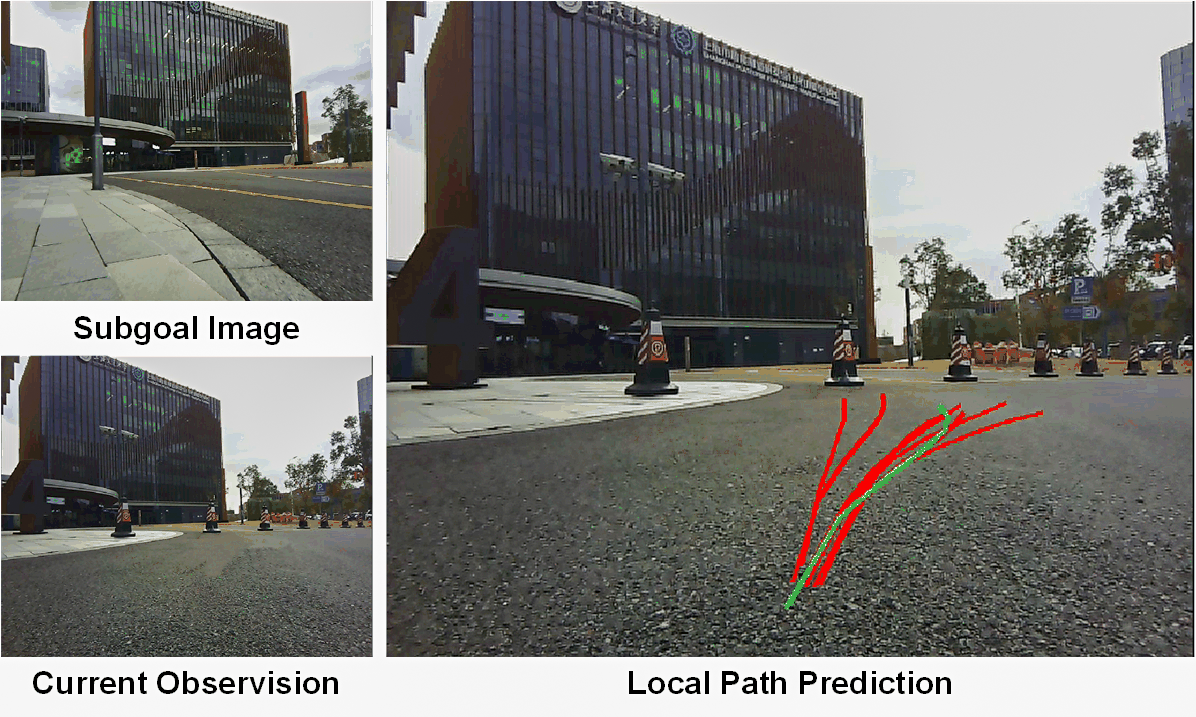}
  \caption{
  Visualization of local path prediction. LP generates multiple local paths based on the current observation and subgoal image, of which GP selects the green path.}
  \label{LP}
  %LP根据当前视觉观察与子目标点图像生成多条局部路径，其中绿色的路径由GP选择
\vspace{-15pt} %缩减空白
\end{figure}

\textbf{Diffusion-Based Motion Generation:} We employ a Denoising Diffusion Probabilistic Model (DDPM)\cite{ho2020denoising} to generate diverse motion candidates. The diffusion process iteratively refines Gaussian noise into viable trajectories through:
%基于扩散的运动生成： 我们采用去噪扩散概率模型（DDPM）来生成不同的运动候选。扩散过程通过以下方式将高斯噪声迭代细化为可行的轨迹：

\vspace{-7pt} % 缩减空间

\begin{equation}
\mathcal{A}_{t-1} = \frac{1}{\sqrt{\alpha_t}} \left(
\mathcal{A}_t - \frac{\beta_t}{\sqrt{1-\bar{\alpha}_t}} \cdot n_{pred}
\right) + \sigma_t \mathbf{z}
\end{equation}
where \( \mathcal{A}_t \) is the noisy trajectory at the current timestep \( t \), initialized as Gaussian noise \( \mathcal{A}_T \sim \mathcal{N}(0, \mathbf{I}) \); \( n_{\text{pred}} \) is the predicted noise by the noise prediction network, which takes as input the noisy trajectory \( \mathcal{A}_t \), the current timestep \( t \), and the multi-modal features including the visual observation and the robot's localization coordinates; \( \alpha_t, \beta_t, \bar{\alpha}_t \) are scheduling parameters computed by the DDPM; \( \sigma_t \mathbf{z} \) is a random noise term added to maintain diversity, with \( \mathbf{z} \sim \mathcal{N}(0, \mathbf{I}) \); and \( \mathcal{A}_{t-1} \) is the updated trajectory after denoising, which iteratively converges to the final trajectory \( \mathcal{A}_0 \) after \( T \) steps.

The noise scheduler performs 50 denoising iterations to produce 8 candidate trajectories $\{a^{(k)}\}_{k=1}^8$, as shown in Figure \ref{LP}.
%each containing 10 waypoints.
%噪声调度器执行 50 次去噪迭代以产生 8 个候选轨迹,如图2所示。

\subsection{Global Planner}
\begin{figure}[thpb]
    \centering
    \begin{subfigure}[b]{0.23\textwidth}
        \centering
        \includegraphics[width=\textwidth]{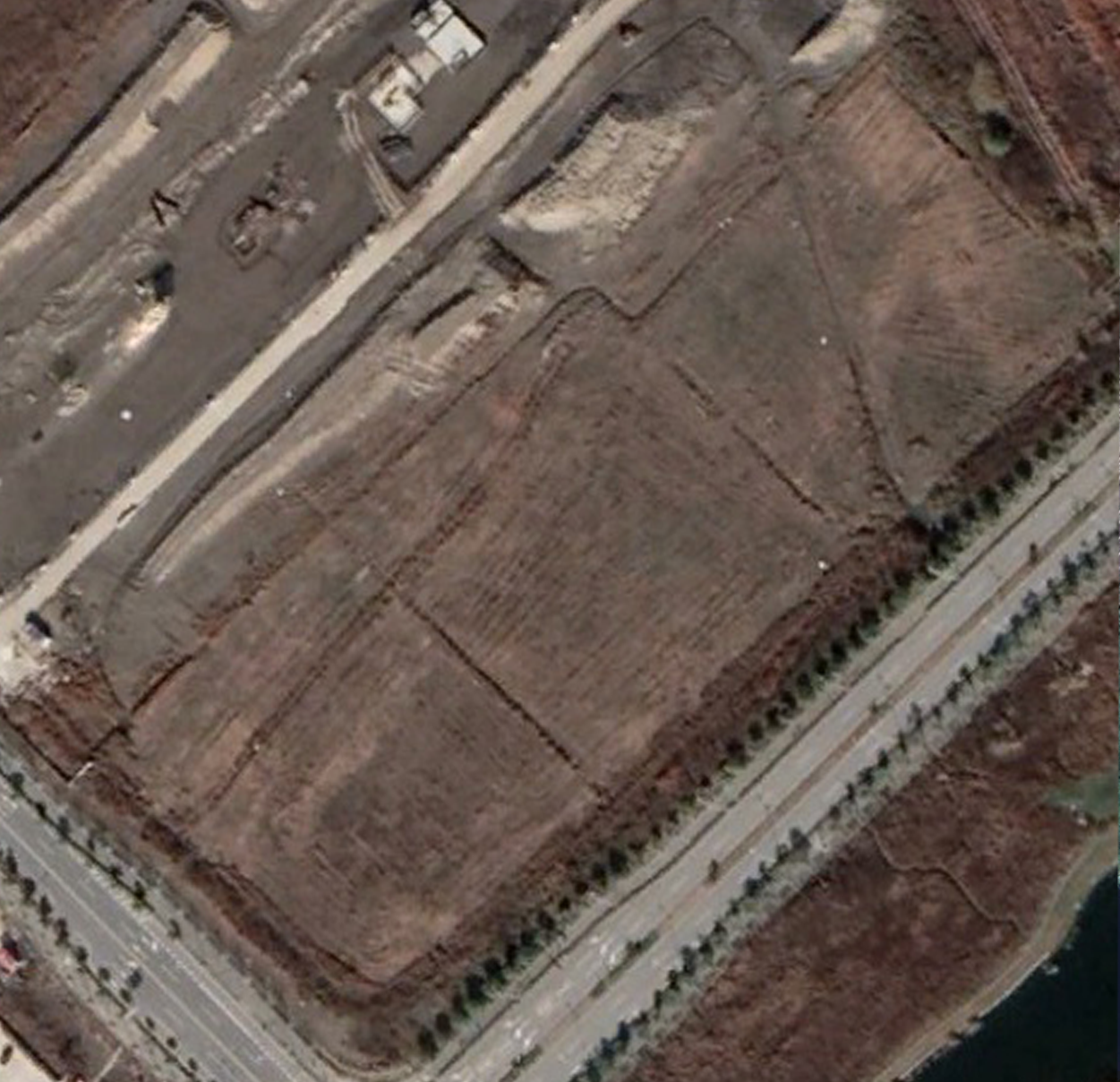}
        \caption{Satellite map}
    \end{subfigure}
    \hspace{2mm}
    \begin{subfigure}[b]{0.23\textwidth}
        \centering
        \includegraphics[width=\textwidth]{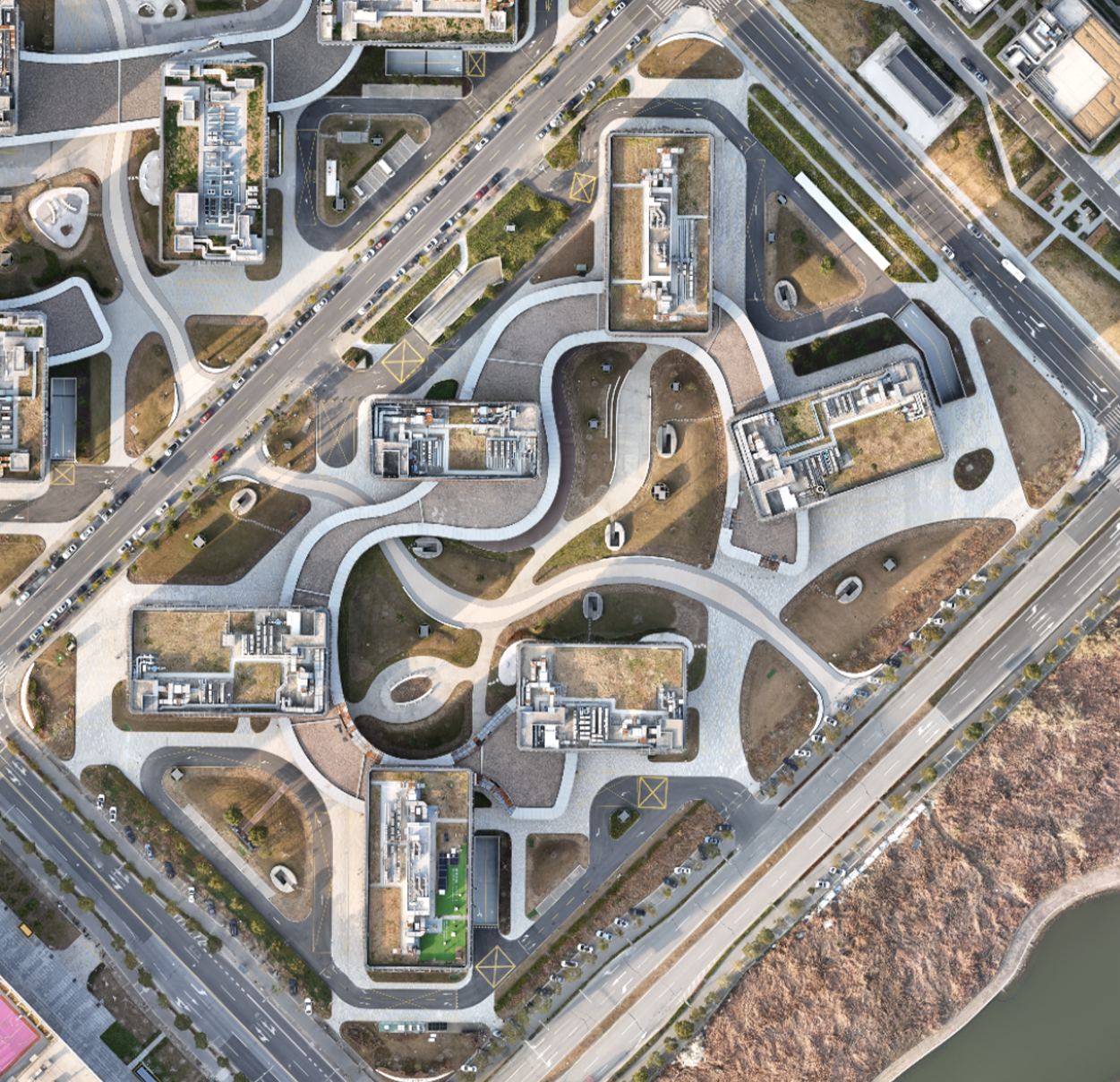}
        \caption{Our global UAV map}
    \end{subfigure}
    \caption{Global UAV map(b) provides updated land cover information compared to satellite map(a) in the study area}
    \label{GP}
\vspace{-10pt} %缩减空白
\end{figure}

While local trajectory optimization ensures short-term safety, global path planning is crucial for efficient and reliable navigation. Our Global Planner (GP) relies on UAV-based Digital Orthophoto Map (DOM), which provide high-resolution environmental awareness and frequent updates. Traditional satellite imagery suffers from low resolution, delayed updates, and occlusions from tall structures, leading to missing or distorted geographic information. In contrast, DOM offer accurate, up-to-date representations, including newly added buildings and ground features. As shown in figure \ref{GP}, the DOM clearly displays buildings absent in the satellite image. This precise mapping is essential for global path planning, enabling reliable navigation based on current environmental data.
%虽然局部轨迹优化确保了短期安全，但全球路径规划对于高效可靠的导航至关重要。我们的全球规划器（GP）依赖于基于无人机的数字正射影像地图（DOM），该地图提供高分辨率的环境感知和频繁的更新。传统的卫星图像遭受低分辨率、延迟更新和高层建筑的遮挡，导致地理信息丢失或扭曲。相比之下，DOM 提供准确的、最新的表示，包括新添加的建筑物和地面特征。如图 \ref {GP} 所示，DOM 清楚地显示了卫星图像中不存在的建筑物。这种精确的映射对于全球路径规划至关重要，能够根据当前的环境数据实现可靠的导航。

Furthermore, in unstructured outdoor environments where landmarks are sparsely distributed with high visual similarity across extended areas, maintaining adherence to the optimal path of VLP Inference becomes particularly challenging. While the LP excels at immediate obstacle negotiation, its limited perceptual horizon (typically$\textless5m$) risks cumulative deviations from the global route over long traversals ($\textgreater 100m$).The introduction of the GP can solve these problems.
%此外，在非结构化的室外环境中，地标分布稀疏，视觉相似性高，跨越扩展区域，保持遵守视觉语言处理器的最佳路径变得特别具有挑战性。虽然本地规划者擅长即时障碍协商，但其有限的感知视野（通常<5m）有可能在长时间穿越（>100m）中累积偏离全球路线。引入全局规划器可以解决这些问题

To extract traversable information from the global map to choose the optimal local path. We have developed the GP.
%为了从全局地图中提取可通过的信息来选择最优的局部路径.我们开发了全局规划器GP

GP generates a probability map that represents the global traversability of the environment on the global map. This probability map helps GP in selecting the optimal path among the candidate local paths. 
%GP在全局地图上生成代表环境全局可穿越性的概率图。该概率图帮助GP在候选局部路径中选择最佳路径。

The U-Net architecture is capable of capturing the spatial details of fine grains through an encoder-decoder structure with skip connections, which we employ to process UAV DOM and output a pixel probability map indicating the traversability of each region in the environment. At the same time, we connect high-dimensional features extracted from the image with trajectory coordinates to generate a single probability score via MLP.
The candidate paths $a^{(k)}$ generated by LP are discretized into $N$ waypoint coordinates by GP, and the traversability score of the paths is calculated. The scoring process is formalized as follows:
% U-Net架构能够通过具有跳过连接的编码器 - 解码器结构捕获细粒度的空间细节，使用它处理自上而下的视图图像，并输出指示环境中每个区域可遍历性的像素概率图。GP使用轨迹数据来学习预测可通行性同时，我们将从图像中提取的高维特征与轨迹坐标连接，通过多层感知机（MLP）产生单个概率分数
%GP将局部规划器生成的候选路径离散为N个航点坐标，从而计算路径的可行性得分
\vspace{-5pt} %缩减空白
\begin{equation}
\text{Score}(a^{(k)}) = \sum_{i=1}^{N} P_m(x_i, y_i) \cdot P_w(x_i, y_i)
\end{equation}
where $(x_i, y_i)$ are the coordinates of the waypoints in the candidate path $a^{(k)}$, $P_m(x_i, y_i)$ is the probability value at each waypoint from the probability map, and $P_w(x_i, y_i)$ is a weighting factor that accounts for the distance and orientation of the waypoints relative to the goal.
% 其中$（x_i，y_i）$是候选路径$a^{（k)}$中航点的坐标, $P_m（x_i，y_i）$是概率图中每个航点的概率值，$P_w（x_i，y_i）$是考虑航点相对于目标的距离和方向的加权因子。

GP uses trajectory data to learn to predict traversability. 
%GP 使用轨迹数据来学习预测可遍历性。
\vspace{-3pt} %缩减空白
\begin{equation}
\mathcal{L} = -\lambda \cdot (1-p_i)^\gamma \cdot \log(p_i)
\end{equation}
where $p_i$ is the predicted probability, the parameter $\lambda$ acts
as a factor, while $\gamma$ adjusts the emphasis on well-classified
examples and hard-misclassified examples. It has been proved by sufficient experiments, that when $\lambda = 3$ and $\gamma = 2$, the network achieves the best training performance.
%其中$p_i$是预测概率，参数$\lambda$作为一个因子，而$\gamma$调整了对分类良好的强调.充分的实验证明，当$\lambda = 3$和$\gamma = 2$时，网络达到了最佳的训练性能。

The GP provides a global perspective on the environment's traversability, guiding the LP to select paths that are both safe and efficient. As shown in Figure \ref{LP}, relying on the GP's global reasoning ability and the timely updated map data of the global map, the robot chooses the optimal path that can bypass the red road post. By leveraging the global map, GP ensures the robot can navigate complex outdoor environments while avoiding untraversable. regions.
%Global Planner（GP）对环境的可遍历性提供全局分析，指导 Local Planner 选择安全高效的路径。如图2所示，依赖于GP的全局推理能力以及全局地图及时更新的地图数据，机器人选择了能够绕过红色路桩的最优路径。通过利用全局地图，GP确保机器人能够在复杂的室外环境中导航，同时避开不可遍历的区域。

\section{Experiments}
This section introduces the experimental setups and presents the results from evaluating the proposed navigation method.
The experiments were conducted in both structured and unstructured environments to assess the performance of our method across diverse conditions. 
The performance of the proposed method is compared against several baseline methods, ViNT\cite{shah2023vint}, GNM\cite{shah2023gnm}, and NoMaD\cite{sridhar2024nomad}, by integrating a large language navigation module into each. 
This integration allows for an evaluation of the navigation performance based on identical language-driven navigation instructions.

\subsection{Test Environments}
The proposed navigation method was evaluated in two distinct outdoor environments: a structured park and an unstructured park. The structured park, featuring paved paths, trees, benches, and light poles, allowed for the assessment of the navigation capabilities of our method. In contrast, the unstructured park, with uneven terrain, natural obstacles such as bushes, and less predictable conditions, presented a more challenging environment. In both settings, dynamic obstacles, including pedestrians and vehicles, were introduced to simulate real-world conditions, requiring the robot to adapt its path-planning and obstacle-avoidance strategies in real time.

\begin{figure*}[thpb]
  \centering
  \includegraphics[scale=0.07]{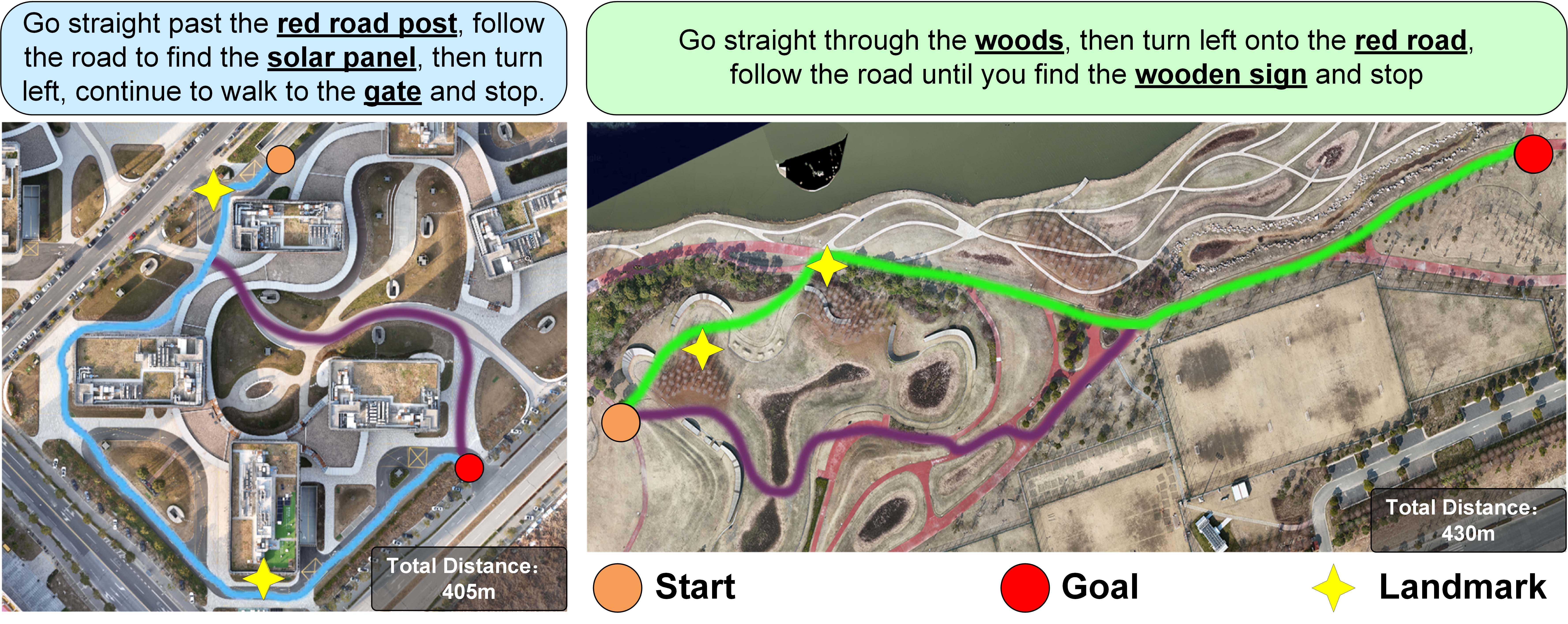}
  \caption{Language-driven instructions guide navigation tasks in structured (left) and unstructured (right) environments. Key landmarks marked with stars include the red road post, solar panel, gate, and other reference points.}
   \label{map}
\end{figure*}

\subsection{Experimental Design}
The robot uses an RGB camera for perception and a GPS receiver for geospatial data. The Jetson AGX Orin handles real-time sensor processing, path planning, and movement control, integrated through ROS for efficient communication.

In each environment, the robot was tasked with completing a 405-meter navigation task in the structured environment and a 430-meter task in the unstructured environment to evaluate its path-planning and obstacle-avoidance capabilities. Data was collected from both environments to create datasets for each setting. Each method was evaluated over 20 trials to ensure statistical reliability and account for potential performance variability. The experiments were conducted under dynamic conditions, with moving obstacles introduced during the navigation task. This setup evaluated the ability of our method to adapt to environmental changes while maintaining path execution efficiency. Figure \ref{map} illustrates the navigation tasks, where the robot follows language-driven instructions to reach the goal. 
% The figure highlights key landmarks such as the red road post, solar panel, and trash can, guiding the robot through its path.
In both environments, the robot selects its path based on the instructions it receives, with the chosen path shown in blue and green.

The baseline methods evaluated in this study include ViNT, GNM, and NoMaD. Since these methods did not initially incorporate a language-driven navigation capability, a vision-language navigation module was integrated into each approach. This allowed each method to be evaluated under the same conditions, utilizing identical language-driven navigation instructions. 
% The inclusion of the language navigation module enabled a comparison of the impact of incorporating language-driven navigation on the overall performance of these baseline methods.

To benchmark the performance of our method, we compared it against several baseline methods: ViNT, GNM, and NoMaD, each representing different approaches to navigation. ViNT focuses on vision-based path optimization, GNM utilizes general navigation policies for robust adaptability across platforms, and NoMaD excels in local path planning in dynamic environments.

We conducted ablation studies to assess the contributions of key components. In Ablation 1 (Table \ref{example}: Ours (LP)), we excluded the GP, using only the LP. In Ablation 2 (Table \ref{example}: Ours (GP)), we removed the LP and relied solely on the GP. These studies helped isolate the effects of the LP and the GP on performance of complete method.
\subsection{Evaluation Metrics}

The performance of the method was evaluated based on the following key metrics, which were selected to comprehensively assess its reliability, efficiency, and adaptability:

\subsubsection{Intervention Count (IC)}

This metric represents the average number of interventions per trial, where intervention occurs when the robot encounters difficulties preventing autonomous recovery, requiring manual assistance, or causing deviation from the intended navigation path. The Intervention Count is calculated as follows:

\vspace{-3pt} % 缩减空白
\begin{equation}
\text{IC} = \frac{\text{Total Number of Interventions}}{\text{Number of Trials}}
\end{equation}

\subsubsection{Path Efficiency (PE)}

This metric evaluates the average path efficiency by comparing the actual path length traveled to the optimal path length, defined as the shortest, obstacle-free path from start to goal. The formula for PE is:
\vspace{-3pt} % 缩减空白
\begin{equation}
\text{PE} = \frac{\text{Actual Path Length}}{\text{Optimal Path Length}}
\end{equation}

\subsubsection{Execution Time (ET)}
This metric measures the average time taken by the robot to complete a specific path, evaluating the time efficiency of the navigation methods. The formula for ET is:
\vspace{-5pt} % 缩减空白
\begin{equation}
\text{ET} = \text{Time to Complete Task (seconds)}
\end{equation}

\subsection{Statistical Analysis}
To validate the significance of the observed differences between the methods, we conducted paired t-tests on IC, PE and ET for each method. The null hypothesis assumed that there was no difference in performance between the methods. These tests were performed in both structured and ET unstructured environments to assess the statistical significance of the observed improvements.

Additionally, $95$\% confidence intervals were calculated for the differences in performance metrics between our method and the baseline methods. The results of these statistical tests are summarized in Table \ref{table2}, which presents the analysis of the performance of the methods.
\subsection{Experimental Results}
The experimental results for IC, PE, and ET for each method in both the structured and unstructured environments are summarized in the following table. Our complete method KiteRunner is compared with ViNT, GNM, NoMaD, and the two ablation studies.

\begin{table}[H]
\setlength{\tabcolsep}{6pt} % Adjust column spacing, smaller values mean narrower spacing
\setlength{\extrarowheight}{2pt} 
\caption{Comparison of Navigation Performance Across Different Methods.}
\label{example}
\begin{tabular}{ccccccccc}

\hline
                         & \multicolumn{3}{c}{Structured}                                                                                                                                                      & \multicolumn{3}{c}{Unstructured}                                                                                                                                                    \\ \cline{2-7} 
\multirow{-2}{*}{Method} & IC                                                        & PE                                                        & ET                                                          & IC                                                        & PE                                                        & ET                                                          \\ \hline
ViNT                     & 13.25                                                     & 1.21                                                      & 718.64                                                      & 20.15                                                     & 1.26                                                      & 862.37                                                      \\
GNM                      & 11.05                                                     & 1.15                                                      & 673.52                                                      & 18.25                                                     & 1.23                                                      & 835.91                                                      \\
NoMaD                    & 6.10                                                      & 1.07                                                      & 628.47                                                      & 15.05                                                     & 1.17                                                      & 792.56                                                      \\
Ours(LP)                 & 5.05                                                      & 1.03                                                      & 615.28                                                      & 12.10                                                     & 1.05                                                      & 738.44                                                      \\
Ours(GP)                 & 5.65                                                      & 1.12                                                      & 654.19                                                      & 15.60                                                     & 1.20                                                      & 817.35                                                      \\
\rowcolor[HTML]{DDE9C5} 
\textbf{Ours(LP+GP)}     & \multicolumn{1}{l}{\cellcolor[HTML]{DDE9C5}\textbf{1.25}} & \multicolumn{1}{l}{\cellcolor[HTML]{DDE9C5}\textbf{1.01}} & \multicolumn{1}{l}{\cellcolor[HTML]{DDE9C5}\textbf{598.12}} & \multicolumn{1}{l}{\cellcolor[HTML]{DDE9C5}\textbf{4.05}} & \multicolumn{1}{l}{\cellcolor[HTML]{DDE9C5}\textbf{1.02}} & \multicolumn{1}{l}{\cellcolor[HTML]{DDE9C5}\textbf{650.35}} \\ \hline
\end{tabular}
\end{table}

\subsection{Statistical Significance}
The statistical significance of performance differences was analyzed using the Wilcoxon signed-rank test, a non-parametric method suitable for non-normally distributed data. The test statistic is defined as:
\vspace{-3pt} % 缩减空白
\begin{equation}
W = \sum_{i=1}^{N} \text{sgn}(x_{i} - y_{i}) \cdot R_i,
\end{equation}

where $x_i$ and $y_i$ are paired observations (e.g., SR values of two methods in the same trial), $R_i$ is the rank of $|x_i - y_i|$, and $N = 20$ is the number of trials. A significance level of $\alpha = 0.05$ was adopted, with the null hypothesis ($H_0$) rejected if $p < \alpha$.

\begin{table}[H]
\setlength{\tabcolsep}{6pt} % Adjust column spacing
\setlength{\extrarowheight}{2pt} 

\caption{Statistical significance (p-values) comparisons.}
\label{table2}
\begin{tabular}{ccccc}
\hline
\multirow{2}{*}{Compared Methods} & \multicolumn{2}{c}{Structured}                        & \multicolumn{2}{c}{Unstructured}              \\ \cline{2-5} 
                                  & IC                        & PE                        & IC            & PE  \\ \hline
ViNT vs Ours (LP+GP)              & \textless{}0.001 & \textless{}0.001 & \textless{}0.001 & 0.001             \\
GNM vs Ours (LP+GP)               & \textless{}0.001 & 0.001 & 0.001 & 0.002  \\
NoMaD vs Ours (LP+GP)             & 0.003 & 0.008 & 0.004 & 0.012             \\
Ours(LP) vs Ours (LP+GP)          & 0.017 & 0.032 & 0.028 & 0.049             \\
Ours(GP) vs Ours (LP+GP)          & 0.002 & 0.003 & 0.005 & 0.007            \\ \hline
\end{tabular}
\end{table}

\subsection{Analysis of Results}

The experimental results in Table \ref{example} highlight the superior performance of the complete method (LP+GP) compared to baseline methods (ViNT, GNM, and NoMaD) across all key metrics: IC, PE, and ET.

In the navigation experiments, manual interventions were performed when the robot deviated from the planned path. The baseline methods exhibited a higher intervention count (IC) in both structured and unstructured environments. In the structured environment, the best-performing baseline method recorded an IC of $6.10$ due to the lack of global information, which made effective long-distance navigation challenging. In the unstructured environment, the lowest IC among baseline methods reached $15.05$ as the sparsity of distinctive landmarks and the high similarity of surrounding scenes hindered effective long-distance navigation, leading to more frequent interventions. In contrast, KiteRunner demonstrated a substantial reduction in intervention count, achieving an IC of $1.25$ in the structured environment, approximately $79.5$\% lower than the baseline, and $4.05$ in the unstructured environment, a reduction of approximately $73.1$\%. These results highlight the superior robustness of KiteRunner, particularly in dynamic scenarios.

In the structured environment, the KiteRunner achieved a near-optimal PE value of 1.01, demonstrating a $5.6$\%–$16.5$\% improvement over baseline methods, with a $5.6$\% gain compared to the best-performing baseline, NoMaD ($1.07$). In the unstructured environment, the KiteRunner attained a PE value of $1.02$, outperforming baselines by $12.8$\%–$19.0$\%, with a $12.8$\% improvement over NoMaD ($1.17$), the highest-performing baseline.

In terms of ET, the KiteRunner completed tasks faster, with times of $598.12$ seconds (structured) and $650.35$ seconds (unstructured), compared to baseline values of $718.64$–$862.37$ seconds. This demonstrates the superior efficiency of our method in both path planning and execution.

Statistical analysis, as shown in Table \ref{table2}, confirms the KiteRunner’s significant performance improvements across all metrics (p $\le$ 0.05). Notably, comparisons with ViNT revealed significant differences in PE and ET (p $\le$ 0.005). Ablation studies further demonstrate the importance of both local and global planners, as removing either component led to substantial performance degradation.

In summary, the KiteRunner significantly enhances navigation efficiency, robustness, and autonomy, confirming its effectiveness in dynamic, real-world environments.

\section{CONCLUSIONS}

This paper presents a novel autonomous navigation method that integrates vision-language processing, UAV orthophoto-based global planning, and diffusion model-driven local trajectory optimization, to achieve efficient and robust navigation in both structured and unstructured environments. 
The proposed method significantly outperforms baselines such as GNM, NoMaD, and ViNT in key metrics, including IC, PE, ET, particularly excelling in dynamic and unpredictable scenarios. 
Its robustness stems from the synergistic combination of global planning for long-range goal-directed navigation and local optimization for real-time responsiveness. 
Ablation studies confirm the critical role of integrating global and local strategies for optimal performance. 
Future work will focus on enhancing real-time adaptability, particularly through UAV-ground collaboration, to improve situational awareness and scalability in large-scale, complex environments. 
The proposed method demonstrates significant potential for advancing autonomous navigation in real-world applications.
%
% \end{thebibliography}
\bibliographystyle{ieeetr} % 指定参考文献样式
\bibliography{ref} % 加载参考文献文件，无需写扩展名

\addtolength{\textheight}{-12cm}   % This command serves to balance the column lengths

\end{document}